\pdfoutput=1
\documentclass[11pt]{article}
\usepackage{authblk}

\usepackage{acl}

\usepackage{times}
\usepackage{latexsym}
\usepackage{colortbl}

\usepackage{cleveref}
\crefname{section}{\S}{\S}
\crefname{Section}{\S}{\S}
\crefformat{section}{§#2#1#3}
\crefname{table}{Tab.}{Tab.}
\crefname{appendix_table}{Tab.}{Tab.}
\crefname{Table}{Tab.}{Tab.}
\crefname{Figure}{Fig.}{Fig.}
\crefname{figure}{Fig.}{Fig.}
\crefname{appendix}{Appendix}{Appendix}
\crefname{chapter}{Chapter}{Chapter}

\usepackage{vcell}
\usepackage{arydshln}
\usepackage{array}
\usepackage{tabularray}
\UseTblrLibrary{booktabs}
\usepackage{tabularx} % to modify spacing of rows and columns
\definecolor{pastel_yellow}{rgb}{0.988,0.898,0.804}
\definecolor{pastel_green}{rgb}{0.784,0.961,0.941}
\definecolor{pastel_purple}{rgb}{0.851,0.824,0.914}
\definecolor{pastel_red}{rgb}{0.902,0.722,0.686}
\definecolor{pastel_blue}{rgb}{0.808,0.851,0.957}
\SetTblrInner{rowsep=0.25pt}
\usepackage{ragged2e} % to control text alignment
\usepackage[T1]{fontenc}
\usepackage[utf8]{inputenc}
\usepackage{graphicx}
\usepackage{subcaption}   

\usepackage{microtype}
\usepackage{multirow}
\usepackage{amssymb}
\usepackage{pifont}
\newcommand{\cmark}{\ding{51}}%
\newcommand{\xmark}{\ding{55}}%
\usepackage{float}

\title{SemEval-2023 Task 10: Explainable Detection of Online Sexism}

\author[1,2*]{\textbf{Hannah Rose Kirk}}
\author[1,3*]{\textbf{Wenjie Yin}}
\author[1]{\textbf{Bertie Vidgen}}
\author[1,2]{\textbf{Paul Röttger}}

\affil[1]{Rewire}
\affil[2]{University of Oxford}
\affil[3]{Queen Mary University of London}

\begin{document}
\maketitle
\def\thefootnote{*}\footnotetext{Equal contribution}\def\thefootnote{\arabic{footnote}}

%%%%%%%%%%%%%%%%%%%%%%%%%%%%%%%%%%%%%%%%%%%%%%%%%%%%%%%%%%%%%%%%%

\begin{abstract}
Online sexism is a widespread and harmful phenomenon. 
Automated tools can assist the detection of sexism at scale.
Binary detection, however, disregards the diversity of sexist content, and fails to provide clear explanations for why something is sexist.
To address this issue, we introduce SemEval Task 10 on the Explainable Detection of Online Sexism (EDOS).
We make three main contributions:
i)~a novel hierarchical taxonomy of sexist content, which includes granular vectors of sexism to aid explainability;
ii)~a new dataset of 20,000 social media comments with fine-grained labels, along with larger unlabelled datasets for model adaptation; and
iii)~baseline models as well as an analysis of the methods, results and errors for participant submissions to our task.
\end{abstract}

\noindent
\footnotesize{\textcolor{red}{\textbf{Content warning:} We show illustrative examples of sexist language to describe the taxonomy and analyse error types.}}
\normalsize
\section{Introduction}
Online sexism can inflict harm on women who are targeted, make online spaces inaccessible and unwelcoming, and perpetuate social asymmetries and injustices. Automated tools are now widely deployed to find sexist content at scale, supporting content moderation, monitoring and research. However, most automated classifiers do not give further explanations beyond generic, high-level categories such as `toxicity', `abuse' or `sexism'. Opaque classification without explanations can cause confusion, anger and mistrust among moderated users, and make it harder to challenge moderation mistakes;
moderators who use automated tools may not trust or fully understand the labels, reducing their efficiency; and it is difficult to assess model weaknesses in the absence of granular labels, which hinders the development of better models.

We organised the SemEval 2023 Task \textit{Explainable Detection of Online Sexism (EDOS)}, which follows from success of previous shared tasks on abuse and hate detection \cite{zampieriSemEval2019, ribeiroINFHatEval2019, pavlopoulosSemeval20212021, basileSemeval2019, fersiniIberEval2018, fersiniEvalita2018}.
Our task proposes and applies a taxonomy with three hierarchical tasks for detecting sexist content. \textbf{Task A} is a binary task to detect whether content is sexist. For sexist content, \textbf{Task B} distinguishes between four distinct categories of sexism, and \textbf{Task C} identifies one of 11 fine-grained sexism vectors.

There are three factors that make our EDOS task unique:
First, \textbf{data diversity}: we sample data from two large social media platforms (Reddit and Gab) using an ensemble of filtering methods. The entries are more diverse than sampling from keywords alone or a single platform.
Second, \textbf{annotation quality}: we recruit highly-trained annotators who all self-identify as women. We make this decision to mitigate implicit biases in labelling and to evoke a participatory approach to AI where the communities primarily harmed from specific forms of content are included in the data development process \cite{birhanePower2022, zytkoParticipatory2022}. To improve consistency of labels with our detailed guidelines, all annotators were trained in multiple pilot tasks, and disagreements resolved by experts.
Third, \textbf{task granularity}: we ground our dataset in a carefully-constructed taxonomy such that the multiple categories in Tasks B and C are both meaningful from a social scientific perspective, aid explainability in predictions and present a challenging machine learning task for multi-class classification.

In releasing the dataset and analysing the accompanying SemEval task submissions, we contribute to efforts in building automated sexism detection systems which are both more \textit{accurate} and \textit{explainable} via fine-grained labels.\footnote{All task data, individual annotations and guidelines are available at \href{https://github.com/rewire-online/edos}{github.com/rewire-online/edos}. A data statement is provided in \cref{sec:data_statement}.}
These contributions mark a step towards a safer internet for women.
\section{Taxonomy Design}
\subsection{Existing Taxonomies}
We conducted a broad review of research on sexism and misogyny. Within this literature, we identified recent articles that propose taxonomies of sexist or misogynist content, in particular 
\citet{jhaWhen2017, samoryCall2021, farrellExploring2019, zeinertAnnotating2021, guestExpert2021, rodriguez-sanchezOverview2021} and \citet{parikhCategorizing2021}.
Each taxonomy is unique, but they generally differ along several dimensions, including differences in (i) construction (whether they are built from theory or empirics); (ii) scope (how the top-level category of sexism or misogyny is defined and which subcategories are included); and (iii) structure (whether subcategories are hierarchically ordered). We review these details in \cref{sec:taxonomy_extra}.

\begin{table*}[h]
\centering
\scriptsize
  \begin{tblr}{
    colspec = {|X[2.25,l]X[3,l]X[7,l]X[4, l]|},
    row{1} = {c,b},
    rows={valign=m},
    column{1} = {font=\bfseries},
  } 
  \toprule
  \textbf{Category}  & \textbf{Vector}     & \textbf{Definition} & \textbf{Example} \\
  \hline
  \SetRow{bg=pastel_yellow}
 \SetCell[r=2]{m}\textbf{1. Threats, plans to harm and incitement} & \textbf{1.1} Threats of harm  & Expressing intent, willingness or desire to harm an individual woman or group of women. This could include, but is not limited to: physical, sexual, emotional or privacy-based forms of harm.  & \textit{I'll kill any women that talks back to me}   \\
 \hline
 \SetRow{bg=pastel_yellow}
& \textbf{1.2} Incitement and encouragement of harm & Inciting or encouraging an individual, group, or general audience to harm a woman or group of women. It includes language where the author seeks to rationalise and/or justify harming women to another person. & \textit{Raping her would put her in her place}\\
\hline  \hline
 \SetRow{bg=pastel_green}
\SetCell[r=3]{m}\textbf{2. Derogation} & \textbf{2.1} Descriptive attacks & Characterising or describing women in a derogatory manner. This could include, but not limited to: negative generalisations about women’s abilities, appearance, sexual behaviour, intellect, character, or morals.  & \textit{Women's football is so shit, they're so slow and clumsy}\\ \hline
 \SetRow{bg=pastel_green}
  & \textbf{2.2} Aggressive and emotive attacks     & Expressing strong negative sentiment against women, such as disgust or hatred. This can be through direct description of the speaker’s subjective emotions, baseless accusations, or the use of gendered slurs, gender-based profanities and gender-based insults. & \textit{I hate women}\\  \hline
 \SetRow{bg=pastel_green}
  & \textbf{2.3} Dehumanising attacks and overt sexual objectification & Derogating women by comparing them to non-human entities such as vermin, disease or refuse, or overtly reducing them to sexual objects. & \textit{Women are pigs}     \\  \hline
  \hline
\SetRow{bg=pastel_purple}
\SetCell[r=4]{m}\textbf{3. Animosity}  & \textbf{3.1} Causal use of gendered slurs, profanities and insults & Using gendered slurs, gender-based profanities and insults, but not to intentionally attack women.  Only terms that traditionally describe women are in scope (e.g. `b*tch', `sl*t').   & \textit{Stop being such a little bitch}\\  \hline
\SetRow{bg=pastel_purple}
  & \textbf{3.2} Immutable gender differences and gender stereotypes  & Asserting immutable, natural or otherwise essential differences between men and women. In some cases, this could be in the form of using women’s traits to attack men. Most sexist jokes will fall into this category. & \textit{Men and women’s brains are wired different bro, that’s just how it is}    \\  \hline
\SetRow{bg=pastel_purple}
  & \textbf{3.3} Backhanded gendered compliments    & Ostensibly complimenting women, but actually belittling or implying their inferiority. This could include, but is not limited to: reduction of women’s value to their attractiveness or sexual desirability, or implication that women are innately frail, helpless or weak.    & \textit{Women are delicate flowers who need to be cherished}    \\  \hline
\SetRow{bg=pastel_purple}
  & \textbf{3.4} Condescending explanations or unwelcome advice & Offering unsolicited or patronising advice to women on topics and issues they know more about (known as `mansplaining').   & \textit{My gf always complains about period pains but she just doesn't understand the medical science for eliminating them!} \\  \hline
  \hline
\SetRow{bg=pastel_red}
\SetCell[r=2]{m}\textbf{4. Prejudiced Discussion}   & \textbf{4.1} Supporting mistreatment of individual women   & Expressing support for mistreatment of women as individuals. Support can be shown by denying, understating, or seeking to justify such mistreatment.      & \textit{Women shouldnt show that much skin, it’s their own fault if they get raped}      \\  \hline
\SetRow{bg=pastel_red}
  & \textbf{4.2} Supporting systemic discrimination against women as a group & Expressing support for systemic discrimination of women as a group. Support can be shown by denying, understating, or seeking to justify such discrimination.    & \textit{The leadership of men in boardrooms is a necessary evil—corporations need to be efficiently run} \\
\bottomrule
\end{tblr}
\caption{Taxonomy of sexism categories (Task B) and fine-grained vectors (Task C).}
\label{tab:vectors_tab}
\end{table*}

\subsection{Our Taxonomy}
\label{sec:our_taxonomy}
Starting from this literature review, we created a first draft of the taxonomy. The taxonomy was further refined using a grounded theory approach \cite{glaserDiscovery2017} with empirical entries from our dataset (see \cref{sec:dataset_construction}) to merge or adjust the schema. It comprises three subtasks:

\paragraph{Task A: Binary Sexism} The first level of our taxonomy makes a binary distinction between sexist and non-sexist content. We define \textit{sexist content} as any abuse, implicit or explicit, that is directed towards women based on their gender, or on the combination of their gender with one or more other identity attributes (e.g. Black women or Muslim women). Our taxonomy focuses on \textit{sexism}, rather than \textit{misogyny}. Misogyny refers to ``expressions of hate towards women" \citep{ussherMisogyny2016}, while sexism also covers more subtle implicit forms of abuse and prejudice that can still substantially harm women.\footnote{Abuse primarily targeted at groups other than women, like racism or antisemitism, is not covered by our taxonomy. However, we tasked annotators with flagging such content using an "other target" label, which, by majority vote, applied to 2,098 out of 15,146 non-sexist entries.}

\paragraph{Task B: Category of Sexism} The second level of our taxonomy disaggregates sexist content into four conceptually and analytically distinct categories. We consciously chose not to separate out categories by the supposed effect on the recipient, or supposed motivation of the speaker, because the harm caused by sexist content is idiosyncratic; and speaker intent is difficult to gauge, especially without broader context. The four categories are \textbf{(1) Threats, plans to harm \& incitement:} Language where an individual expresses intent and/or encourages others to take action against women which inflicts or incites serious harm and violence against them. It includes threats of physical, sexual or privacy harm. \textbf{(2) Derogation:} Language which explicitly derogates, dehumanises, demeans or insults women. It includes negative descriptions and stereotypes about women, objectification of their bodies, strong negative emotive statements, and dehumanising comparisons. It covers negative statements directed at a specific women and women in general. \textbf{(3) Animosity:} Language which expresses implicit or subtle sexism, stereotypes or descriptive statements. It includes benevolent sexism, i.e., framed as a compliment.
\textbf{(4) Prejudiced Discussion:} Language which denies the existence of discrimination, and justifies sexist treatment. It includes denial and justification of gender inequality, excusing women's mistreatment, and the ideology of male victimhood.

%%%%%%%%%%%%%%%%%%%%%%%%%%%%%%%%%%%%%%%%%%%%%%%%%%%%%%%%%%%%%%%%%%
\paragraph{Task C: Fine-Grained Vectors} The third level of our taxonomy disaggregates each \textit{category} of sexism into fine-grained sexism \textit{vectors}. We seek vectors which are mutually exclusive (i.e., each vector is distinct) and collectively exhaustive (i.e., all sexist content can be assigned to a vector). \cref{tab:vectors_tab} gives an overview of the 11 fine-grained vectors in our taxonomy, along with their definitions and examples taken from prior literature then edited to clearly demonstrate each distinct vector.

\section{Dataset Construction}
\label{sec:dataset_construction}

\subsection{Choice of Platform}
A large portion of online harms research gathers data from Twitter \cite{vidgenDirections2020}, resulting in lacking community-based diversity in the severity and form of sexist content. Thus, in order to produce a diverse dataset with coverage of our fine-grained vectors of sexism, we opt to use two social media platforms. First, \textbf{Gab}, which is an `alt-tech' social networking site established in 2016. It has positioned itself as a rival to mainstream sites, such as Twitter and Facebook, and explicitly aims to protect free (and far-right) speech \cite{zannettouWhat2018,limaRightLeaning2018, jasserWelcome2021}. Second, \textbf{Reddit}, which is a network of topic-based forums (`subreddits'), where users can share, view and comment on content related to their interests. Several 
studies have shown that Reddit is home to many sexist and anti-feminist communities, described as the `manosphere' \cite{gingAlphas2017, zuckerbergNot2018, farrellExploring2019, gingNeologising2020, ribeiroEvolution2021}. 

We source equal amounts of data from Reddit and Gab. For each platform, we first create a pool of 1M entries (see \cref{sec:data_collection}).  We then sample 10k entries from each pool for labelling (see \cref{sec:data_filtering}).

\subsection{Data Collection}
\label{sec:data_collection}
\paragraph{Gab}
We collect 34M publicly available Gab posts from August 2016 to October 2018.\footnote{\href{https://files.pushshift.io/gab/}{files.pushshift.io/gab}}
This data has been widely used in academic studies \citep[e.g.,][]{kennedyGab2018, cinelliEcho2021}. We randomly sample 1M entries to create the pool.

\paragraph{Reddit}
First, we compile a list of 81 subreddits which are likely to contain sexist content, based on previous work \cite{guestExpert2021, farrellExploring2019, zuckerbergNot2018, jonesSluts2020, qianBenchmark2019, gingAlphas2017, ribeiroEvolution2021} and online wikis.\footnote{\href{https://rationalwiki.org/wiki/}{rationalwiki.org/wiki} and \href{https://incels.wiki/}{incels.wiki}} Then, we collect all comments from August 2016 to October 2018 in these subreddits using the Reddit API.\footnote{We only collect comments as our early analysis showed that many posts are single URLs, images or videos and thus harder to parse in isolation for labelling. We specify the date range to match the Gab data.}
We manually review each subreddit and assign it to one of four categories (Incels, Men Going Their Own Way, Men’s Rights Activists, Pick Up Artists) based on prior work by \citet{lillyWorld2016}.\footnote{We describe these categories in \cref{sec:additional_reddit_detail}.}
To ensure that our dataset is not overly biased towards niche linguistic expressions and topics, we restrict our sampling to the 24 subreddits with at least 100k comments, resulting in a dataset of 42M comments.
Finally, we randomly sample 250k comments from each of the four subreddit categories to create the pool.

\begin{table*}[h]
\scriptsize
\centering
\begin{tblr}{
  colspec = {X[12,l] X[0.7,c]X[0.7,c]|X[0.7, c]X[0.7,c]|X[0.7,c]X[0.7, c]},
  row{1} = {c,b, font = \bfseries},
  rows={valign=m},
  }
  \toprule
  & \SetCell[c=2]{c}train & & \SetCell[c=2]{c}dev & & \SetCell[c=2]{c} test &\\
\hline
\SetRow{bg=pastel_blue}
\SetCell[c=7]{l}\textbf{Task A} & & & & & & \\
\hline
not sexist  & 10,602 &76\%   & 1,514 &76\%  & 3,030 &76\%    \\
sexist     & 3,398 & 24\%    & 486 &24\%   & 970 &24\%     \\ 
\cline{2-7}
\SetCell[c=1]{r}\textit{total}  & 14,000 &100\%  & 2,000 &100\% & 4,000 &100\%   \\ 
\hline
\SetRow{bg=pastel_blue}
\SetCell[c=7]{l}\textbf{Task B} & & & & & & \\
\hline
1. threats, plans to harm and incitement    & 310 &9\% & 44 &9\%     & 89 &9\%  \\
2. derogation   & 1,590 &47\%    & 227 &47\%   & 454 &47\%     \\
3. animosity    & 1,165 &34\%    & 167 &34\%   & 333 &34\%     \\
4. prejudiced discussion & 333 &10\%     & 48 &10\%    & 94 &10\% \\ 
\cline{2-7}
\SetCell[c=1]{r}\textit{total}  & 3,398 &100\%   & 486 &100\%  & 970 &100\%    \\ 
\hline
\SetRow{bg=pastel_blue}
\SetCell[c=7]{l}\textbf{Task C} & & & & & & \\
\hline
1.1 threats of harm  & 56 &2\%  & 8 &2\% & 16 &2\%  \\
1.2 incitement and encouragement of harm    & 254 &7\% & 36 &7\%     & 73 &8\%  \\
2.1 descriptive attacks   & 717 &21\%     & 102 &21\%   & 205 &21\%     \\
2.2 aggressive and emotive attacks     & 673 &20\%     & 96 &20\%    & 192 &20\%     \\
2.3 dehumanising attacks and overt sexual objectification      & 200 &6\% & 29 &6\%     & 57 &6\%  \\
3.1 casual use of gendered slurs, profanities, and insults & 637 &19\%     & 91 &19\%    & 182 &19\%     \\
3.2 immutable gender differences and gender stereotypes    & 417 &12\%     & 60 &12\%    & 119 &12\%     \\
3.3 backhanded gendered compliments    & 64 &2\%  & 9 &2\% & 18 &2\%  \\
3.4 condescending explanations or unwelcome advice    & 47 &1\%  & 7 &1\% & 14 &1\%  \\
4.1 supporting mistreatment of individual women  & 75 &2\%  & 11 &2\%     & 21 &2\%  \\
4.2 supporting systemic discrimination against women as a group & 258 &8\% & 37 &8\%     & 73 &8\%  \\ 
\cline{2-7}
\SetCell[c=1]{r}\textit{total}  & 3,398 &100\%   & 486 &100\%  & 970 &100\%    \\
\bottomrule
\end{tblr}
\caption{Distribution of class labels across tasks and data splits.}
\label{tab:data_split_labels}
\end{table*}

\subsection{Data Preparation and Sampling}
\label{sec:data_filtering}
\paragraph{Cleaning} We cleaned the text in the Gab and Reddit pools by: (1) replacing URLs and usernames with generic tokens; (2) dropping empty entries; (3) dropping entries that only contain URLs or emoji; (4) dropping non-English language entries; and (5) dropping duplicates. After these cleaning steps were completed, we apply our sampling techniques.

\paragraph{Boosted Sampling} The prevalence of abusive content `in the wild' is difficult to estimate reliably, but could be as low as 0.1\% or 1\% \citep{vidgenChallenges2019}. Sexism represents only a subset of all abuse; So, random sampling from online platforms will lead to a dataset with a large class imbalance, which impedes the training of AI systems.
A range of sampling techniques have been used in prior work to boost the proportion of abusive content in datasets, including keyword search \cite{elsheriefNotOkay2017} or lexicons \cite{farrellExploring2019}, and community-based \cite{guestExpert2021, vidgenIntroducing2021} or user-based sampling \cite{vidgenChallenges2019}. For privacy concerns, we do not store user-based information. Instead, we apply a mix of community-based sampling (on Reddit), with a carefully-designed ensemble of varied sampling methods.

\paragraph{Ensemble of Sampling Methods} Relying on a single sampling method makes the sampled data more prone to biases \cite{yin2022hidden}. Following extensive pilots, we settle on six different techniques to sample 10k cleaned entries from each of Gab and Reddit pools (total $n=20{,}000$), to ensure coverage of the 11 fine-grained sexism vectors and introduce lexical and topical diversity. The six techniques sample (1) 1,000 entries with at least one sexist keyword (e.g. ``c*nt'', ``b*tch''); (2) 1,000 entries with at least one sexist keyword and one topical keyword (e.g. ``she'', ``girl''); (3) 100 entries from each decile of toxicity scores from the Perspective model (total 1,000)\footnote{\href{https://www.perspectiveapi.com/}{perspectiveapi.com}}; (4) 100 entries from each decile of scores from a bespoke classifier for sexism detection which we trained on seven open source sexism/misogyny datasets (total 1,000)\footnote{For training details, see \cref{sec:bespoke_classifier}.}; (5) 1,000 entries from cases where the score from Perspective's Toxicity model differed from the score of our custom classifier by at least $|0.8|$; and (6) 5,000 entries sampled using a combination of topical keywords and scores from other attributes of Perspective (e.g., Identity Attacks + ``girl'' or Sexually Explicit + ``she'').

\subsection{Data Annotation}
We follow the `prescriptive paradigm' for data annotation, in that we want the annotators to apply our comprehensive annotation guidelines rather than applying their own subjective beliefs \cite{rottgerTwo2022a}. Our annotation guidelines contain definitions, clarifying notes, exemplars, edge cases, and general guidance.

\paragraph{Annotator Recruitment}
To mitigate the risk of implicit bias and to encourage participation from affected communities, we worked with trained annotators who all self-identify as women.
We required that all annotators pass a challenging 200-entry screening task that covered all 11 sexism vectors in our data.
In total, we recruited and trained 19 annotators who passed the screening.
We opted for expert annotation over crowdwork because pilot experiments demonstrated an marked difference in quality and consistency of labels.\footnote{Results are presented in \cref{sec:crowd_experiment}.} The demographics of annotators and their experiences with online sexism are documented in \cref{sec:data_statement}.

\paragraph{Annotation Process}
Three annotators labelled each entry. To further ensure label quality, we rely on expert adjudication for disagreements. Experts were called upon to give labels for (i) cases with less than 3/3 agreement (unanimous) in Task A, and (ii) cases with less than 2/3 agreement in Tasks B and C. The expert team consisted of two women authors of this paper, and two of the most experienced workers from the annotation team. Data was assigned to annotators in batches every two weeks over the course of two months. Throughout the process, we worked closely with annotators so that their feedback could be incorporated into our guidelines and so that their welfare could be continuously monitored and protected.

\subsection{Dataset Distribution}
The data was split into training, development, and test sets in the ratio of 70:10:20. Only sexist instances are included in Task B and C. Label distributions of all splits are shown in \cref{tab:data_split_labels}.

\section{Task Description}
\subsection{Task Definition}
Our SemEval task consists of three subtasks, reflecting our hierarchical taxonomy (\cref{sec:our_taxonomy}). Task A is a binary classification task (sexist vs. not sexist). Task B and C are multi-class classification tasks, with four and 11 categories, respectively.

\subsection{Task Organisation}
We ran our SemEval task on CodaLab.\footnote{\href{https://codalab.lisn.upsaclay.fr/competitions/7124}{codalab.lisn.upsaclay.fr/competitions/7124}} There were two primary phases: (i) the Training and Development Phase which ran from September 2022 to January 2023, and (ii) the Test Phase, which began on 10th January 2023 and ended on 31st January 2023. During the Training and Development Phases, we released entries and labels from the train and dev splits, as well as an additional \textit{starting kit} including 1M unlabelled Reddit entries and 1M unlabelled Gab entries. During the Test Phase, we staggered the release of test entries for Task A from Tasks B and C, since the latter provides information on the correct labels of the former. 

\subsection{Evaluation Metrics and Baselines}
\label{sec:baselines}
\begin{table*}[t!]
\scriptsize
\centering
\begin{tblr}{
  colspec = {X[3,r]X[5,c]X[2, c]X[1,c]X[1,c]X[1, c]},
  row{1,2,3} = {c,b, font = \bfseries},
  } 
  % top row
  \toprule
  \SetRow{bg=pastel_yellow}
  \SetCell[c=6]{c} Baselines \\
  \hline
  & & Continued  & \SetCell[c=3]{c}Macro-F1 \\
 & Model & Pre-training & Task A & Task B & Task C \\
 \hline
 \textit{B0} & MostFrequent & \xmark & 0.4310 & 0.1594 & 0.0317  \\
\textit{B1}& Uniform & \xmark  & 0.4509 & 0.2413 & 0.0629  \\
\textit{B2}& XGBoost & \xmark  & 0.4933 & 0.2297 & 0.0881  \\ 
\hline
\textit{B3}& \SetCell[r=2]{c}DistilBERT  & \xmark & 0.7621 & 0.5531 & 0.2935  \\
\textit{B4}& & \cmark  &  0.7804 &  0.5367 &  0.3140 \\
\hline
\textit{B5}& \SetCell[r=2]{c}{DeBERTa-v3-base}  & \xmark  & 0.8235 & 0.4790  & 0.1517  \\
\textit{B6}& &  \cmark &  \textbf{0.8235} &  \textbf{0.5926} &  \textbf{0.3171}  \\
 \hline \hline
\SetRow{bg=pastel_green}
\SetCell[c=6]{c} \textbf{Top Ranked Systems} \\ \hline
\textit{PingAnLifeInsurance} & DeBERTa-v3-large, twHIN-BERT-large & \cmark  & \textbf{\textcolor{red}{0.8746}} &  &   \\
\textit{stce}   & RoBERTa-large, ELECTRA   & \cmark  & 0.8740& 0.7203& 0.5487 \\
\textit{FiRC-NLP} & DeBERTa (ensemble)& \xmark  & 0.8740&  &    \\
\textit{JUAGE}  & PaLM (ensemble)  & \xmark  & \ & \textbf{\textcolor{red}{0.7326}} &    \\
\textit{PASSTeam} & RoBERTa, HateBERT  & \cmark  &   & 0.7212& 0.5412 \\
\textit{PALI}   & DeBERTa, RoBERTa  & \cmark  &   &   & \textbf{\textcolor{red}{0.5606}}  \\
\hline \hline
\SetRow{bg=pastel_purple}
\SetCell[c=6]{c} \textbf{Summary Statistics}\\ \hline
\SetCell[c=3]{r} \textbf{System count} & &  & 84     & 69     & 63      \\
\SetCell[c=3]{r}\textbf{Q1} & & &  0.7994    & 0.5730    & 0.3153     \\
\SetCell[c=3]{r}\textbf{Mean} & && 0.8095    & 0.5899    & 0.3829     \\
\SetCell[c=3]{r}\textbf{Median} &  &     & 0.8322    & 0.6191    & 0.4230     \\
\SetCell[c=3]{r}\textbf{Std} &  &   & 0.0746    & 0.1065    & 0.1274     \\
\SetCell[c=3]{r}\textbf{Q3} &  &  & 0.8537    & 0.6501    & 0.4758     \\
\SetCell[c=3]{r} \textbf{Percentage of systems which beat \textit{B0}} & &   & 100\% & 100\% & 100\% \\ 
\SetCell[c=3]{r} \textbf{Percentage of systems which beat \textit{B6}} & &   & 55\%   & 60\%   & 73\% \\
\bottomrule
\end{tblr}
\caption{Baselines, top ranked leaderboard results and summary statistics per Task. The best baseline is \textbf{bolded}. The best participant submission is \textbf{\textcolor{red}{bolded in red}}.}
\label{tab:summary_results}
\end{table*}
We evaluate all systems with macro-average F1 score to account for imbalance between classes. We supply 7 baseline models (see \cref{tab:summary_results}) to benchmark a range of simple to complex systems. The simplest baselines are always predicting the most frequent class (\textit{B0}) and uniformly predicting each class (\textit{B1}). We train one traditional machine learning model (\textit{B2}) where the data is first vectorized with TF-IDF and then fitted with an XGBoost model. The remaining baselines use DistilBERT \cite{sanhDistilBERT2019} and DeBERTa-v3-base \cite{heDeBERTaV32021}. For each model, we fine-tune on the training set (\textit{B3, B5}), and do continued pre-training on the 2M unlabelled entries from Reddit and Gab combined (\textit{B4, B6}). Across all tasks, DeBERTa with continued pre-training sets the highest baseline. The best baseline for Task A (F1=0.8235) is more saturated than for Tasks B and C (F1=0.5926; 0.3171), making this an interesting SemEval challenge with varying degrees of difficulty and remaining headroom for performance improvements.

\section{Participant Systems and Results}

\subsection{Participant Overview}
EDOS was a very popular SemEval task, with 599 accounts signing-up for dataset access.\footnote{Number of accounts signed up as of 25/04/2023.}
In the Development Phase, 128 submissions were made for Task A, 71 for Task B and 52 for Task C.
In the Test Phase, 134 submissions were made for Task A, 87 for Task B and 81 for Task C.

\begin{figure*}[t]
    \centering
    \includegraphics[width = \textwidth]{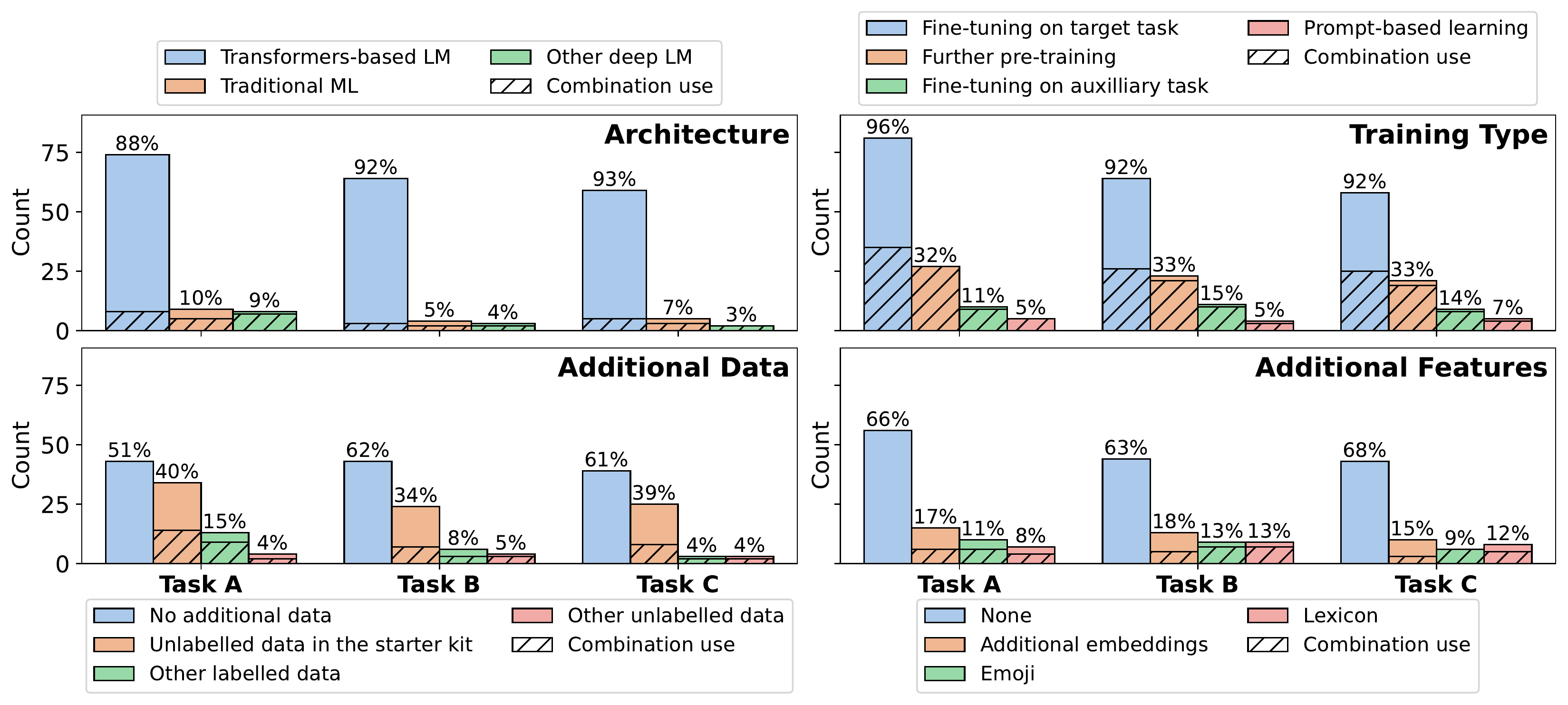}
    \caption{Participant methodologies by popularity (count and percentage of submissions using the method). Each subplot shows a different breakdown of system paper method decisions. The first subplot (top left) shows that the majority of participants used transformer-based language models. The second subplot (top right) shows fine-tuning on the target task was the most popular training method, but a substantial proportion of participants also applied further pre-training. The third subplot (bottom left) similarly shows that while the majority only used the labelled data provided in the task, many participants also used the in-domain unlabelled data in our starter kit. Finally, the last subplot (bottom right) shows the majority did not use additional features.}
    \label{fig:popularity_bar}
\end{figure*}

\paragraph{Validation Process} To ensure a fair competition in the Test Phase, we allowed one submission account per team and up to two test submissions of task predictions, to allow for one upload mistake.\footnote{We unfortunately encountered some suspicious activity with multiple accounts using generic emails \texttt{ABCD1234@domain}, or making >10 test set submissions.} We issued a short survey to confirm compliance with our T\&Cs and collect system information. Our final leaderboard includes those who responded to this survey: 84 to Task A, 69 to Task B and 63 to Task C. 59 teams submitted to all three tasks, 10 submitted to two, and 20 submitted to just one.

\subsection{Leaderboard Results}
The top three submissions for each task are presented in \cref{tab:summary_results}.\footnote{Full leaderboards are available on \href{https://github.com/rewire-online/edos}{github.com/rewire-online/edos}}
The top three systems in all tasks use multiple models or an ensemble, and many top systems apply further pre-training and multi-task learning. Notably, \textit{stce} and \textit{PASSTeam} achieved top results on two or more tasks. Both used multi-task learning and further pre-training on the unlabelled data in the starter kit: \textit{stce} use RoBERTa-large \cite{liuRoBERTa2019} and ELECTRA \cite{clarkELECTRA2020}; while \textit{PASSTeam} use a multi-task learning strategy \cite{yuGradient2020} with fine-tuned RoBERTa and HateBERT \cite{caselliHateBERT2021}.
In Task A, \textit{PingAnLifeInsurance} also adopt a multi-task DNN structure \cite{liuMicrosoft2020} with further pre-training of DeBERTa-v3 \cite{heDeBERTaV32021} and TwHIN-BERT \cite{zhangTwHINBERT2022} on the starter kit unlabelled data and an additional Kaggle dataset. \textit{FiRC-NLP} use an ensemble of various DeBERTa models fine-tuned only on the labelled task data. In Task B, \textit{JUAGE} was one of the few systems relying on prompt-based learning, but achieved first place using an instruction-tuned PaLM model \cite{chowdheryPaLM2022} with a parameter-efficient prompt tuned on the task data and majority voting over six iterations. In Task C, \textit{PALI} further pre-trained DeBERTa-v3 using the starter kit unlabelled data and applied a second loss term (normalized temperature-scaled cross entropy).

Scores for Tasks B and C were substantially lower in mean and higher in variance than Task A (see \cref{tab:summary_results}). All participant systems beat our simplest baseline predicting the most frequent class, while a majority still beat our more complex baseline of DeBERTa-v3 with continued pre-training.

\begin{figure*}
    \centering
    \includegraphics[width = \textwidth]{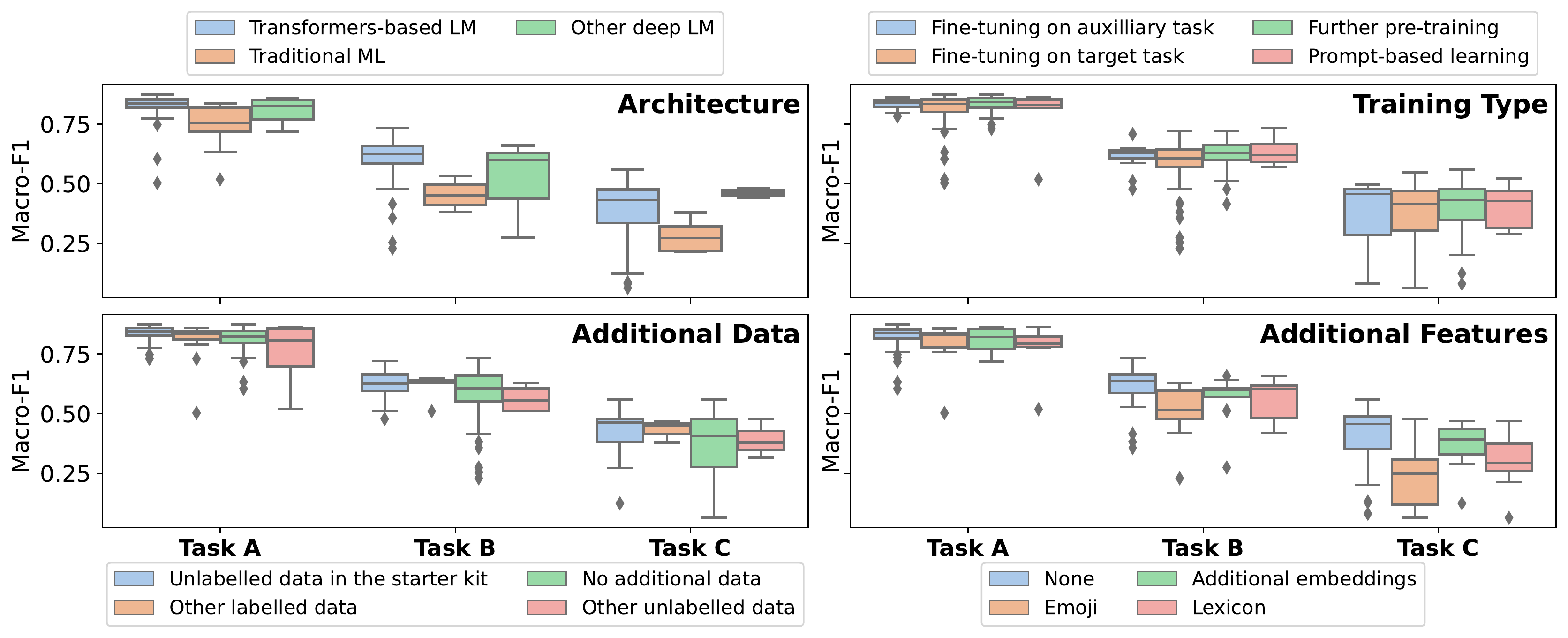}
    \caption{Participant methodologies by Macro-F1 score (across submissions using the method in isolation or combination). Each subplot shows a different breakdown of system paper method decisions and the average F1 score by task. F1 scores on Task A are generally higher and more tightly distributed than on Tasks B and C.}
    \label{fig:performance_box}
\end{figure*}

\subsection{Popular Methods}
\paragraph{Architecture} For all tasks, the large majority of participants ($\sim$90\%) used a transformers architecture (\cref{fig:popularity_bar}). The most popular transformer-based models include RoBERTa \cite{liuRoBERTa2019}, DeBERTa \cite{heDeBERTa2021, heDeBERTaV32021}, BERT \cite{devlinBERT2019}, BERTweet \cite{nguyenBERTweet2020} and DistilBERT \cite{sanhDistilBERT2019}. Several submissions use prompted language models such as GPT-2 \cite{radfordLanguage2019}, GPT-3 \cite{brownLanguage2020a}, PaLM \cite{chowdheryPaLM2022} or OPT \cite{zhangOPT2022a}. Other approaches include traditional machine learning methods ($\sim$8\%) and non-transformer deep neural networks ($\sim$6\%), which are often combined with other methods.

\paragraph{Additional Training and Data}
The majority of participants applied fine-tuning only on the target task (>90\%) but some also apply further pre-training ($\sim$30\%), fine-tuning on an auxiliary task ($\sim$15\%) or prompt-based learning ($\sim$5\%). Around 40\% of participants used the unlabelled data in the starter kit. Participants also sometimes used resources outside our task, both labelled ($\sim$10\%) and unlabelled data ($\sim$5\%). A number of innovative methods were applied including data augmentation, active learning or data cartography \cite{swayamdiptaDataset2020a}.

\paragraph{Additional Features} The majority of participants did not use additional features ($\sim$66\%). If additional features were used, they were predominantly used in a combination of additional embeddings ($\sim$18\%), emoji ($\sim$12\%) or lexicons ($\sim$12\%).

\subsection{Most Effective Methods}
The distribution of participant F1 scores across methodologies is more tightly clustered for Task A than Tasks B and C (see \cref{fig:performance_box}). Across methodology variables, \textbf{architecture} choices lead to the most different distributions of performances, with transformers-based and other deep language models providing a more competitive average F1 than traditional ML systems across tasks. In contrast, \textbf{training type} seems to have the least influence on the distribution of F1 scores, with the most overlap between conditions, even in Task C. The highly related \textbf{additional data} condition has slightly more effect on performance, with the use of unlabelled data in the starter kit resulting in the highest average. Systems that did not use any \textbf{additional features} had higher average F1 scores in all tasks.

\section{Error Analyses}
We base our error analyses on the 10 best performing systems for each task as their errors indicate the remaining difficulty within our dataset.

\subsection{Quantitative Error Analysis}

\paragraph{Confusion Between Binary Labels} Out of 4,000 test instances in Task A, 162 were misclassified by all top 10 systems, where 38.3\% were false positives (not sexist content predicted as sexist) and 61.7\% were false negatives (sexist content predicted as not sexist). Of the false negatives, 41\% are from the \textit{Animosity} category, 34\% from \textit{Derogation}, 16\% from Prejudiced Discussion and 9\% from Threats. This pattern is expected because \textit{Animosity} is often contains implicit language, while \textit{Threats} are commonly conveyed in explicit language.

\begin{figure}[!b]
    \centering
    \includegraphics[width = 0.96\columnwidth]{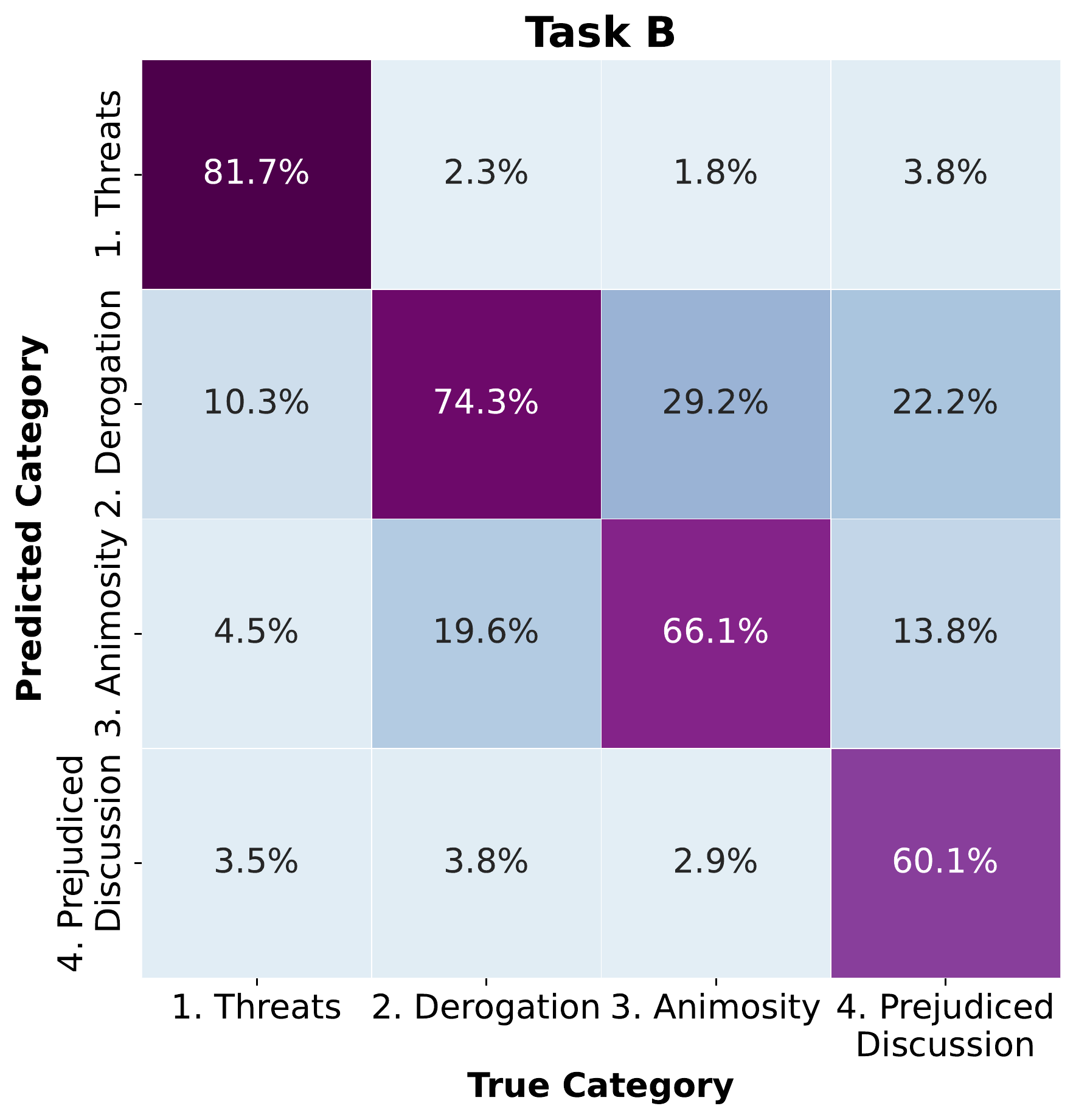}
    \caption{Confusion matrix for categories of sexism (Task B) across the top 10 performing systems.}
\label{fig:cm_task_b}
\end{figure}

\paragraph{Confusion Between Categories and Vectors}
Among the categories, \textit{Threats} is least often misclassified as other categories and the least common mistake for instances with the true label of any other category (see \cref{fig:cm_task_b}). \textit{Animosity} and \textit{Derogation} are often mistaken for each other. Instances of \textit{Prejudiced Discussion} are also commonly misclassified as \textit{Derogation} or \textit{Animosity}. The confusion across vectors in Task C has a similar pattern (\cref{sec:confusion_vectors}). Some errors reflect class imbalance: the largest class (2.1) and the smallest class (3.4) are the most and least frequent confusions, respectively.
Others errors reveal inherent similarity between certain vectors: 2.1 and 3.2 are commonly confused, with both referring to gendered stereotypes, but 2.1 is explicitly negative in sentiment. The same is true for 2.2 and 3.1, which share gendered slurs, except that 3.1 is causal (not targeted) usage.

\subsection{Manual Error Inspection}
\label{sec:manual_error}
Two self-identifying woman authors manually inspected 100 instances from the Task A test set that were misclassified by all top 10 teams. Out of the 100 entries, we identified 76 as genuine system misclassifications, while 24 were annotation errors.\footnote{Note that the annotation error rate is likely higher in this subset relative to the dataset as whole, because it includes challenging examples that both automated systems and humans struggle to classify.}

\paragraph{System Errors} Of the 76 system errors, 52 were false negatives (68\%) and 24 were false positives (32\%). Of the \textbf{false negatives}, there were several common themes of misclassified examples. These themes include (i) expressions that are seemingly positive in sentiment or backhanded gendered compliments (e.g., ``\textit{STEM ain't hard, wamans can do it too!}''); (ii) gender stereotypes encoded in a joke (e.g., ``\textit{I just bought a smart car. It doesn't turn on when the wife is in the driving seat.}''); (iii) slang use (e.g., ``\textit{Tha Brothas want dem PAWGs.}'') or rarer gendered insults (such as ``hag'' or ``witch''); and (iv) irony (e.g., ``\textit{either you worship women or you're a misogynist.}''). Some explicit examples, paired with rhetorical devices such as questions, were also misclassified as not sexist (e.g., ``\textit{How does one learn how to choke a women? I assume there is an art to it? Is it that you're depriving blood to the brain more so than choking the windpipe?}'').

Of the \textbf{false positives}, there were also a number of common patterns. These include (i) counter-speech (e.g., ``\textit{It's fucked up how women aren't allowed to even be neutral. Always have to put on a fucking show to entertain men and keep them from attacking you.}''); (ii) matter-of-fact discussion or descriptions of groups associated with sexism (e.g., ``\textit{The whole point of MGTOW is, men are going their own way....which doesn't mean we've changed from desiring women to desiring men. It just means we don't allow women to decide what it is to be a man - or to be the main focal point of our lives.}''); or (iii) criticisms of an individual woman's actions (e.g., ``\textit{I know a woman who slept with one off her best friends on off boyfriend. Dropped her entire social circle for a jerk, not even a Chad. She was an, 8 as well. SMH. She had choices, chose this guy. Women can be dumb.}'').

\paragraph{Annotation Errors} Out of the 24 mislabelled examples, 17 were false negatives (29\%) and 7 were false positives (71\%). In general, annotators struggled with the distinction between individual-related and generalised statements. False negatives occur when gender stereotypes are expressed as speculations rather than characteristic descriptions (e.g., ``\textit{We have condoms and birth control so now women can sex with fewer and much less serious consequences as well. It makes sense that they would have more sex with the top guys when there's relatively little to risk.}"). False positives occur with non-gendered attacks on individual women targets (e.g., ``\textit{This woman is just stupid.}'').

\paragraph{Grey Areas with Lacking Context} Although a decision was reached by the authors on each manually-inspected instance, 9 of the 24 mislabelled instances were considered a possibly grey area -- instances that could fall under a different label depending on context or slight modifications to the guideline interpretation. Individual posts and comments taken in isolation sometimes lack the context to interpret meaning or intent. Annotators may then fill in the context based on their experiences with online sexism or knowledge of the relevant communities. This was particularly evident in criticisms of individual woman, where it is often unclear without context whether the speaker is criticising an individual for their specific actions or using that individual as scapegoat to express general sexist stereotypes and views. This also raises the question of whether something should be flagged if it hints at the speaker's sexist tendency but does not itself convey specific sexist content.

\section{Discussion}
\subsection{Lessons Learnt}
\paragraph{Value of Diverse Data Sampling}
We note that previous SemEval tasks on hate or abuse detection primarily report false positive errors from their participants. For example, in HatEval \cite{basileSemeval2019}, 89.1\% of the errors made by the top three systems were false positives. In our analysis, we instead find more false negatives. We believe this flipped distribution of error types is due to the subtle and diverse forms of sexist content in our dataset (such as \textit{Prejudiced Discussion}) which do not necessarily share linguistic form or keywords with more explicit sexist content, making them more challenging to correctly identify.

\paragraph{Challenge of Fine-Grained Predictions}
Our baselines and participant scores demonstrate that binary sexism detection (Task A) is a substantially easier task than fine-grained sexism detection (Task B and C). The maximum F1 score achieved in Task C is just above 50\%. This shows that high scores reported in binary detection tasks of socially-conditioned concepts (like sexism, hate or abuse) obscure fine-grained model failings in distinguishing between forms of content, which may have very different impacts on their targets and society at large. Given the remaining headroom in predicting our categories and vectors of sexism, we are excited to see how future model-centric contributions using our dataset continue to improve performance. Future work could also apply data-centric techniques to increase performance and robustness, such as augmenting the dataset with challenging adversarial examples \cite{kiela2021dynabench, vidgenLearning2021, kirkHatemoji2022a} or mapping hard-to-learn subspaces with data cartography methods \cite{swayamdiptaDataset2020a}.

\paragraph{Efficiency of Continued Pretraining} Labelling data for socially-conditioned tasks (like sexism) is challenging, expensive and requires some degree of domain expertise \cite{kirkMore2022}. However, providing unlabelled in-domain data is both cheap, scalable and avoids causing psychological harm on annotators from a greater labelling burden \cite{kirkHandling2022}. We are encouraged that our best baselines and top performing participant systems effectively used continued pre-training on the unlabelled in-domain data that we provided (from Reddit and Gab). We encourage future SemEval tasks to similarly release unlabelled datasets from which they sample smaller labelled datasets.

\subsection{Limitations}
\paragraph{Class Imbalance} Our dataset is more balanced on the binary label (24\% sexism) than many previous datasets for hate and abuse detection \cite{vidgenDirections2020}. However, there is substantial imbalance in categories and vectors of sexism. Thus, it is hard to confirm whether confusion between categories and vectors is due to inherent features of the data or due to the class imbalance. We encourage future work to examine the effect on performance and cross-vector confusion when balancing the dataset. That said, we explicitly made the decision to not re-balance our dataset because different types of sexism (especially at the fine-grained level) do have very different base rates `in the wild'. Dealing with class imbalance is then part-and-parcel of the problem we seek to address. 

\paragraph{Defining ``Explainable''} We cast the explainability problem as a classification task on a single text document (without network or user information). In this setting, combining hierarchical classifications is what adds explainability by demonstrating whether a piece of content is \textit{sexist} and for what reason (\textit{category} and \textit{vector}). There are many other promising methods to make automated systems more explainable for socially-sensitive tasks (e.g., sexism, abuse or hate), including  classifying spans or masked tokens for rationale prediction \cite{mathewHateXplain2021, kimWhy2022}; leveraging user and network data \cite{qianLeveraging2018a, wichExplainable2021} or targets data \cite{kennedyContextualizing2020}; applying social bias frames for stereotypes \cite{sapSocial2020a}; or encouraging multi-hop and chain-of-thought reasoning \cite{zhangRethinking2022, jinWhen2022}.

\paragraph{Value Specificity and Perspective} Our dataset is labelled by UK-residing annotators, who all self-identify as women. We issue prescriptive guidelines, then rely on majority vote and expert adjudication to produce a gold label. Many previous studies reveal that annotator identity is a critical determinant of labelling behaviour \cite{waseemAre2016a,larimoreReconsidering2021,sapAnnotators2022b} and so majority votes inadequately capture subjective disagreement \cite{davaniDealing2021a}. To encourage future work on annotator-specific artefacts, we provide annotator IDs with each labelled entry \cite{prabhakaranReleasing2021a}. Nonetheless, our definition of sexism, its subcategories in our taxonomy and how our annotators apply their best judgement are grounded in few cultural viewpoints, primarily from Western-centric and English-speaking communities. Any classifiers built from our dataset will thus inherit our values (as taxonomy and guideline writers) and those of our annotators.\footnote{ \citet{bangEnabling2022} seek to design ``value-aligned'' sexism detection systems, where their model considers both the content and an individual's value system to return a classification. This technique is a promising way to further improve inclusivity and explainability in sexism detection tasks.} 
Like most toxic content datasets, our data is limited to English \citep{rottger-etal-2022-data}.
The bounds of sexism, however, are culturally contested and people may legitimately disagree in where to draw the line.

\section{Conclusion}
In this paper, we made three main contributions.
First, we provided a new taxonomy for the more explainable classification of sexism in three hierarchical levels -- binary sexism detection, category of sexism and fine-grained vector of sexism. This taxonomy is grounded in reviews of prior taxonomies and social science literature, then empirically-validated with in-domain data from two social media platforms. It thus provides a sociotechnical overview of the varied and nuanced landscape of online sexism.
Second, we created a high-quality dataset, sampled with an ensemble of techniques to increase the diversity of content and annotated by self-identifying women experts to ensure consistent labels. This labelled dataset is paired with a larger unlabelled dataset to mitigate the constraints that a labelling budget (both in terms of financial cost and psychological cost to annotators) has on the efficacy of trained systems.
Finally, we shared baseline models and summarised systems submitted to our SemEval task. This analysis demonstrates the success of techniques that combine state-of-the-art transformer models with continued pre-training and multi-task learning. However, there is still substantial headroom for improving performance on detecting fine-grained forms of sexist content.
We hope that our research and resources can contribute towards the design of future systems that make online spaces safer for all.

\section*{Acknowledgements}
This work was carried out by Rewire, and funded by MetaAI.
We are grateful to our annotators, and to the Rewire team for providing feedback on this article.
We also thank all participants who submitted results to our task, and our reviewers who gave valuable and constructive comments on the system papers.
\section*{Ethical Risks and Harm Statement}
\label{sec:harm_statement}
We release a dataset containing online sexism, where we have demonstrated that state-of-the-art models still err on some specific entries and vector types. Malicious actors could use these fine-grained failures as inspiration for sexist online posts which bypass current detection systems, or in principal use the entries to train a generative model (concerns also shared by \citet{vidgenLearning2021} and \citet{kirkHatemoji2022a}). We believe this risk is manageable given the benefit facilitated by our dataset in understanding and predicting online sexism.

Following \citet{kirkHandling2022}'s advice, we describe the risks from our dataset construction and release. First, there is a risk of harm to data subjects (women targeted by sexist entries) and readers of the paper in reproducing or reinforcing harmful representations, stereotypes, prejudiced discussions or slur usage. We include a content warning on the first page and consistently display quoted examples in italic text (\cref{tab:vectors_tab}, \cref{sec:manual_error}). Where possible, we replace vowels in slurs with an asterisk. Due to the labelling intensiveness of this work (20k entries), there is a risk to annotators from repeatedly viewing sexist content, especially violent entries (like `threats of harm'). We carefully follow protocols for supporting annotator well-being and keep a direct line of communication open to them via a group messaging forum. We survey annotators at the end of the labelling process to understand how the task affected their well-being and how we can do better in the future. Overall, annotators found we adequately supported their mental health and ensured the process was as safe as possible. A few annotators suggested we run a workshop presenting findings of our work. We commit to hosting this de-brief session in the coming months as it is important for annotators to see the significance of their work in supporting online safety.

\newpage
\bibliography{edos_references}
\bibliographystyle{acl_natbib}

\clearpage
\appendix
\section{Data Statement}
\label{sec:data_statement}
We provide a data statement \cite{benderData2018a} to document the generation and provenance of our EDOS dataset.

\subsection{Curation Rationale}
The purpose of the EDOS dataset is to train and evaluate automated systems for the fine-grained and explainable detection of online sexism. We curate social media comments across diverse forms of sexist content -- varying from explicitly threatening behaviour (e.g., ``I'll kill any women that talks to me'') through to more subtle forms (e.g., ``Women are delicate flowers who need to be cherished'').

\subsection{Language Variety}
All entries are in English, dictated by the expertise of researchers and annotators. We encourage future work to expand our vectors to other languages.

\subsection{Speaker Demographics}
All entries are collected from two social media platforms: Gab and Reddit. Thus, the speaker demographics approximate these platforms' users in general \cite{amayaNew2021}. We expect (and intentionally sample) users who are likely male, western and right-leaning, and hold extreme or far-right views about women, gender issues and feminism \cite{zannettouWhat2018, limaRightLeaning2018}.

\subsection{Annotator Demographics}
\label{sec:annotator_demographics}
We recruited 19 annotators from our existing network of freelance annotators. They worked over two months in Spring 2022. Annotators were paid \textsterling16/hour and expert annotators were paid \textsterling20/hour. All annotators were screened on a gold standard set of 200 entries and had previous experience working on hate speech annotation tasks. Annotators could contact the research team at any point using a messaging platform. All annotators self-identified as women. We sent annotators an optional survey to collect further information on their demographics and to understand past experiences with social media and online sexism. Of 19 annotators, 12 responded to our survey. By age, 3 were between 18--24 years old, 8 were between 25--30 years old and 1 was between 31--40 years old. The completed education level was high school for 1 annotator, undergraduate degree for 6 annotators and post-graduate degree for 5 annotators. Annotators came from a variety of nationalities, with 7 British, as well as Swedish, Swiss, Italian and Argentinian. Most annotators identified as White with one annotator identifying as mixed South Asian and White, and one annotator identifying as Black British. The majority identified as heterosexual (7), with others identifying as bisexual, pansexual or asexual. 8 annotators were native English speakers and 4 were non-native but fluent. The majority of annotators used social media for one or more hours a day (7). All annotators had seen others targeted by online sexism, and 7 had been personally targeted. 

We also collected annotator-specific attitudes to online sexism, for example (i) what types of online sexism they believe social media companies should prioritise in their content moderation policies, and (ii) whether content should be removed or partially hidden. We ask annotators about their content moderation preferences for content ascribing to each of our 11 vectors. In future work, we plan to map these preferences to our dataset entries to investigate whether attitudes influenced labelling behaviour.

\subsection{Speech Situation}
The modality of entries is short-form written textual comments from social media, which were interpreted in isolation for labelling i.e., not in a comment thread and without user or network information.

\subsection{Text Characteristics}
The genre of texts is sexist and non-sexist social media comments. The composition of the final dataset is described in \cref{tab:data_split_labels}.

\section{Review of Existing Taxonomies}
\label{sec:taxonomy_extra}
We reviewed recent articles that propose taxonomies of sexist or misogynist content, in particular:  
\citet{jhaWhen2017, samoryCall2021, farrellExploring2019, zeinertAnnotating2021, guestExpert2021, rodriguez-sanchezOverview2021}; and \citet{parikhCategorizing2021}. We describe here several key differences between prior work and how our taxonomy (in \cref{sec:our_taxonomy}) compares:

\paragraph{Differences in Construction} The taxonomies we reviewed are either theoretically- or empirically-grounded.
Empirically-motivated studies mostly produce a first version of the taxonomies based on NLP literature and then iteratively adjust the taxonomy using new data they collect \cite{zeinertAnnotating2021, guestExpert2021}.
Twitter is the most widely used source \cite{jhaWhen2017, fersiniIberEval2018, samoryCall2021, zeinertAnnotating2021, rodriguez-sanchezOverview2021}, followed by Reddit \cite{farrellExploring2019, guestExpert2021}. Data collection was mostly based on keywords, such as slurs and sexist hashtags, except for \citet{farrellExploring2019} who sampled more widely from the ``manosphere" community and \citet{parikhCategorizing2021} who based their study on ``The Everyday Sexism Project" -- a catalogue of accounts of people who experienced sexism. Some studies also included the re-annotation of existing datasets \cite{jhaWhen2017, samoryCall2021, guestExpert2021} and addition of adversarial data \cite{samoryCall2021}. Our taxonomy is theory- and empirics-grounded -- we first draw on previous theory-based taxonomies and then iterate with in-domain data.

\paragraph{Differences in Scope} Existing taxonomies differ in scope, which is partly reflected in their focus on either sexism \cite{samoryCall2021, rodriguez-sanchezOverview2021,parikhCategorizing2021} or misogyny \cite{jhaWhen2017, fersiniIberEval2018, farrellExploring2019, guestExpert2021, zeinertAnnotating2021}.
There are clear inconsistencies in how both of these terms are defined. \citet{zeinertAnnotating2021}, for example, formally define misogyny as a type of group-directed abuse and hate speech. In practice, their account of misogyny is very similar to the fairly broad account of sexism that we have adopted.
Our taxonomy clearly positions sexism as a type of abuse, with a broader scope than misogyny. Existing taxonomies often cover overlapping but different topics. They also differ in how they name and define more fine-grained types of sexism and misogyny -- which we call vectors -- if those are present at all.
\citet{jhaWhen2017}, for instance, use a specific theory on benevolent sexism from social psychology to motivate three main types of sexism: Paternalism, Gender Differentiation, and Heterosexuality.
In contrast, \citet{parikhCategorizing2021} construct 23 vectors of sexism, based on the perspectives of those targeted by sexism.
They distinguish, among other things, between role and attribute stereotyping, as well as work-, menstruation-, and motherhood-related gender discrimination.
Some work uses descriptive names for vectors (e.g., ``physical violence" \cite{farrellExploring2019, guestExpert2021}, while other names are based on social science theories or terms (e.g., ``gaslighting" \citep{parikhCategorizing2021}, ``belittling" \citep{farrellExploring2019} and ``neosexism" \citep{zeinertAnnotating2021}). Our taxonomy consistently uses descriptive names, with a roughly equivalent degree of granularity within each level.

\paragraph {Differences in Structure} Finally, existing taxonomies differ in how they organise vectors of sexist and misogynist content.
Most commonly, all vectors are direct subtypes of misogyny or sexism \cite{fersiniIberEval2018, zeinertAnnotating2021, farrellExploring2019, rodriguez-sanchezOverview2021, parikhCategorizing2021}, without explicit relationships between them.
\citet{jhaWhen2017} and \citet{samoryCall2021} include additional flags for benevolent/civil vs. hostile/uncivil expressions, and \citet{fersiniIberEval2018} have flags for targets. 
\citet{guestExpert2021} provide a multi-level hierarchy of types and vectors of misogyny. Our taxonomy also takes a hierarchical approach, differentiating first between categories of sexism and then between fine-grained sexism vectors within each category.

\section{Additional Detail on Reddit Data}
\label{sec:additional_reddit_detail}
We identified a long list of 81 relevant subreddits by reviewing seven research papers and three established online lists. Some of the subreddits have been banned but data is still available for them, and they are included within our long list. The 81 subreddits were then assigned to one of four categories, based on the work of \citet{lillyWorld2016}.
Lilly's work was also used in \citet{ribeiroEvolution2021}, who adapted it to create a five-part taxonomy of manosphere subreddits by separating `Men's Rights Activists' (MRA) into MRA and `The Red Pill' (TRP). We do not follow this distinction because conceptually the two categories are very similar and separating them could bias our data collection by, in effect, oversampling from these similar communities.
To maximise the diversity of our data we include subreddits which are topically relevant to the four categories (listed below), even if they are nominally dominated by groups other than straight men (such as \texttt{r/gaycel} or \texttt{r/trufemcels}) or are primarily concerned with critiquing and debating the manosphere (such as \texttt{ r/purplepilldebates} or \texttt{r/thebluepill}).
\begin{itemize}
    \item \textbf{Incels (IC)} Men who are opposed to, and demean, insult or attack women, because they cannot, or believe they believe  cannot, get sexual interest or companionship. Many such men believe that they are entitled to female attention.
    \item \textbf{Men Going Their Own Way (MGTOW)} Men who believe that women, and feminism, have corrupted society and that men need to reassert themselves. Many such men can, or believe that they can, achieve companionship and sexual interest from women and are less explicitly spiteful and hateful towards women, in contrast to Incels.
    \item \textbf{Men's Rights Activists (MRA)} Men who typically oppose feminism, believing that women are privileged and  men are systematically discriminated against. MRA groups range from moderate positions, such as activists who want greater rights for divorced fathers, to more extreme positions, such as activists who want men to be given state-backed privileges.
    \item \textbf{Pick Up Artists (PUA)} Men who actively attempt to win companionship and sexual interest from women, often with duplicitous or underhand techniques. Many such men hold derogatory views about women, and portray them as sexual objects.
\end{itemize}

\section{Bespoke Sexism Classifier for Data Sampling}
\label{sec:bespoke_classifier}
In our sampling ensemble, we trained a bespoke binary classifier on English and women-targeted entries from seven open source datasets \cite{zampieriSemEval2019, rodriguez-sanchezOverview2021, vidgenLearning2021, samoryCall2021, guestExpert2021, fersiniIberEval2018, fersiniEvalita2018}.
For each of these datasets, we remove duplicates, clean white space, and convert URLs, emoji and usernames to special tokens (e.g., [URL]).
Each dataset is labelled according to its own taxonomy for sexism and/or misogyny, and their structure and definitions are not consistent.
To unify the labels, we use the binary top-level label in each dataset (either misogynist vs. not misogynist; or sexist vs. not sexist).
We split the combined datasets into train and test (90/10), stratifying by dataset source.
We train a BERT-base model with default parameters for 3 epochs, using the transformers library \cite{wolfTransformers2020a}.
The model achieves 80\% Macro-F1 score on the held-out test set. We use this model to weakly label the presence of misogynist or sexist content in the Gab and Reddit data.

\section{Crowdworkers vs. Experts Experiments}
\label{sec:crowd_experiment}
We used a gold standard of 200 entries to test two alternatives for annotation. First, we launched the gold standard on a crowdsourced annotation service, with 7 annotations per entry, and 176 crowdworkers.
Taking the majority-voted label for Task A, nearly all sexist entries were labelled correctly (96\%) but most non-sexist entries were mislabelled (4\% correct).
Of the 57 Sexist entries, the crowd majority-vote was 28\% correct for Task B and 14\% for Task C.
Second, we recruited and trained 19 self-identifying women to label the same 200 gold standard entries.
Taking the majority-voted label for Task A, the trained annotators were correct 96\% of the time. 
Of the 57 Sexist entries, the trained annotators' majority-vote was 84\% correct for Task B and 72\% for Task C. Given the sensitivity and complexity of the task, working with trained annotators provides higher-quality labels and poses other advantages that annotator welfare can be monitored and guidelines can be updated to address feedback.

\section{Confusion Between Vectors}
\label{sec:confusion_vectors}
\vspace{-1em}
\begin{figure}[H]
    \centering
    \includegraphics[width = 0.87\columnwidth]{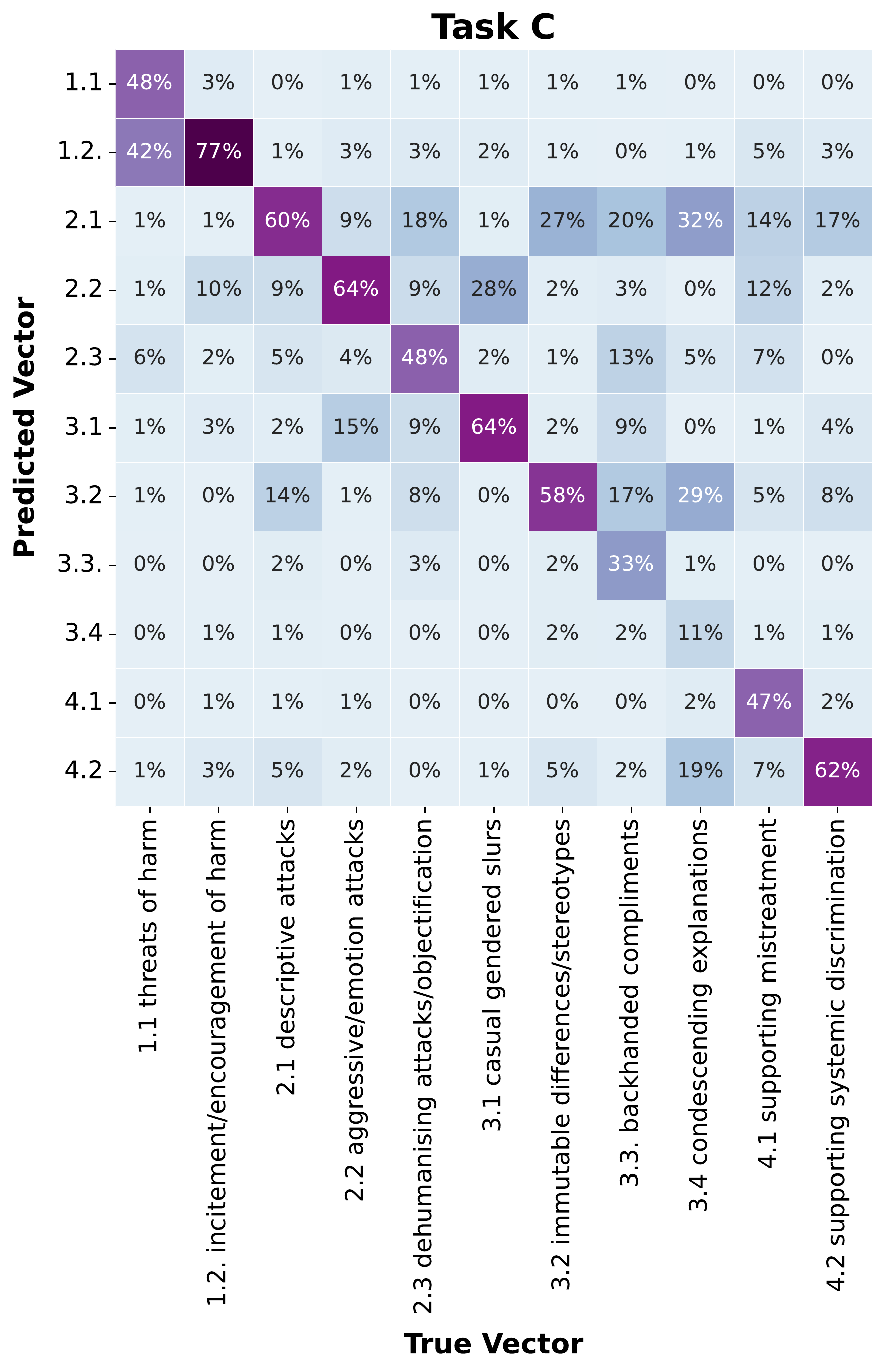}
    \caption{Confusion matrix for vectors of sexism (Task C) across the top 10 performing systems.}
    \label{fig:my_label}
\end{figure}
\vspace{-2em}

\end{document}